%% file: conference_101719.tex
\documentclass[conference]{IEEEtran}
\IEEEoverridecommandlockouts
\usepackage{cite}
\usepackage{amsmath,amssymb,amsfonts}
\usepackage{algorithmic}
\usepackage{graphicx}
\usepackage{textcomp}
\usepackage{xcolor}
\usepackage{url}
\usepackage{wasysym}
\def\BibTeX{{\rm B\kern-.05em{\sc i\kern-.025em b}\kern-.08em
		T\kern-.1667em\lower.7ex\hbox{E}\kern-.125emX}}
\newcommand{\pixel}{\text{pixel}}
\newcommand{\unchecked}{$\times$}
\begin{document}
	
	\title{
		Feature Boosting, Suppression, and Diversification for Fine-Grained Visual Classification
	}
	
	\author{\IEEEauthorblockN{Jianwei Song, Ruoyu Yang}
		\IEEEauthorblockA{\textit{National Key Laboratory for Novel Software Technology} \\
			\textit{Nanjing University, Nanjing 210023, China}\\\
			songjianwei@smail.nju.edu.cn, yangry@nju.edu.cn}
	}
	
	\maketitle
	
	\begin{abstract}
		Learning feature representation from discriminative local regions plays a key role in
		fine-grained visual classification. Employing attention mechanisms to extract part features has
		become a trend. However, there are two major limitations in these methods:
		First, they often focus on the most salient part while
		neglecting other inconspicuous but distinguishable parts. Second, they
		treat different part features in isolation while neglecting their relationships. To handle these limitations, we propose to locate multiple different
		distinguishable parts and explore their relationships in an explicit way. In this
		pursuit, we introduce two lightweight modules that can be
		easily plugged into existing convolutional neural networks.
		On one hand, we introduce a feature boosting and suppression module that boosts the most salient part of feature maps
		to obtain a part-specific representation and suppresses it to force the following network to mine other potential parts. On the other hand, we introduce a feature diversification module that learns semantically complementary information from the correlated part-specific representations.
		Our method does not need bounding boxes/part annotations
		and can be trained end-to-end. Extensive experimental results
		show that our method achieves state-of-the-art performances
		on several benchmark fine-grained datasets. Source code is available at \url{https://github.com/chaomaer/FBSD}.
	\end{abstract}
	
	\section{Introduction}
\begin{figure}[htbp]
		\centerline{\includegraphics[width=0.8\linewidth]{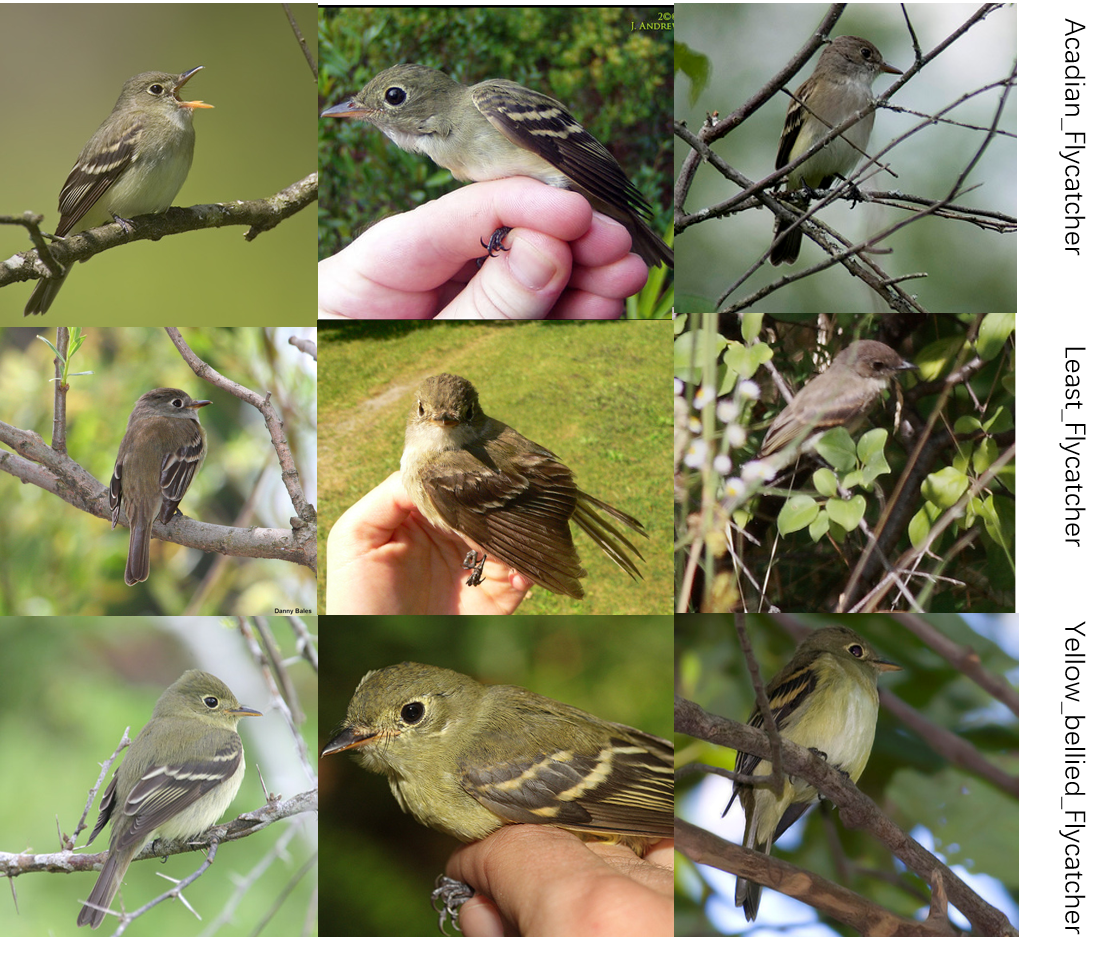}}
		\caption{Illustration of large intra-class and small inter-class variations in FGVC. The images with large variations in each row belong to the same class. However, the images with small variations in each column belong to different classes. This situation is opposite to generic visual classification.}
		\label{fig:fgvc}
	\end{figure}

	Fine-grained visual classification (FGVC) focuses on distinguishing subtle visual
	differences within a basic-level category, e.g., species of birds\cite{CUB} and dogs\cite{DOG}, and models of aircrafts\cite{CRAFT} and cars\cite{CAR}.
	Recently, convolutional neural networks (CNNs) have made great progress on many tasks, such as face recognition\cite{facenet}, automatic driving \cite{self-drive},
	Pedestrian re-identification \cite{Part-Pool}, intelligent logistics in IOT and so on.
	However, traditional CNNs are not powerful enough to capture the subtle discriminative features due to the large intra-class and small inter-class variations as shown in Fig. \ref{fig:fgvc}, which makes FGVC still a challenging task. Therefore, how to make CNNs locate the distinguishable parts and learn discriminative features are important issues that need to be addressed.
	Early works \cite{Part-RCNN}\cite{Deep-LAC}\cite{Part-SCNN}\cite{Pose-Norm} relied on predefined bounding boxes and part annotations to capture visual differences. However, collecting extra annotated information is labor-intensive and requires professional knowledge, which makes these methods less practical. Hence, researchers recently have focused more on weakly-supervised FGVC that only needs image labels as supervision. There are two paradigms towards this direction. One is based on part features, these methods  \cite{RA-CNN}\cite{FDL}\cite{NTS}\cite{MGE}\cite{S3N} are often composed of two different subnetworks. Specifically, a localization subnetwork with attention mechanisms is designed for locating discriminative parts and a classification subnetwork is followed for recognition. The dedicated loss functions are designed to optimize both subnetworks. The limitation of these methods is that it is difficult to optimize because of the specially designed attention modules and loss functions.
	The other is  based on high-order information,
	these methods \cite{BCNN}\cite{Compact-BCNN}\cite{iSQRT-COV}\cite{HCNN}\cite{Low-Rank} argue that the first-order information is not sufficient to model the differences and instead use high-order information to encode the discrimination. The limitation of these methods is that it takes up a lot of GPU resources and has poor interpretability.
	
	We propose feature boosting, suppression, and diversification towards both efficiency and interpretability. We argue that attention-based methods tend to focus on the most salient part, so other inconspicuous but distinguishable parts have no chance to stand out \cite{GA-CNN}. However, the network will be forced to mine other potential parts when masking or suppressing the most salient part.
	Based on this simple and effective idea, we
	introduce a feature boosting and suppression module (FBSM), which highlights the most salient part of feature maps at the current stage to obtain a part-specific representation and suppresses it to force the following stage to mine other potential parts.
	By inserting FBSMs into the middle layers of CNNs, we can get multiple part-specific feature representations that are explicitly concentrated on different object parts.
	
	Intuitively, individual part-specific feature representation neglects
	the knowledge from the entire object and may not see the forest for the trees. To eliminate the bias, we introduce a feature diversification module (FDM) to diversify each part-specific feature representation. Specifically, given a part-specific representation, we enhance it by aggregating complementary information discovered from other parts. Through modeling the part interaction with FDM, we make the part-specific feature representation more discriminative and rich. 
	
\begin{figure*}
		\centering
		\includegraphics[width=\linewidth]{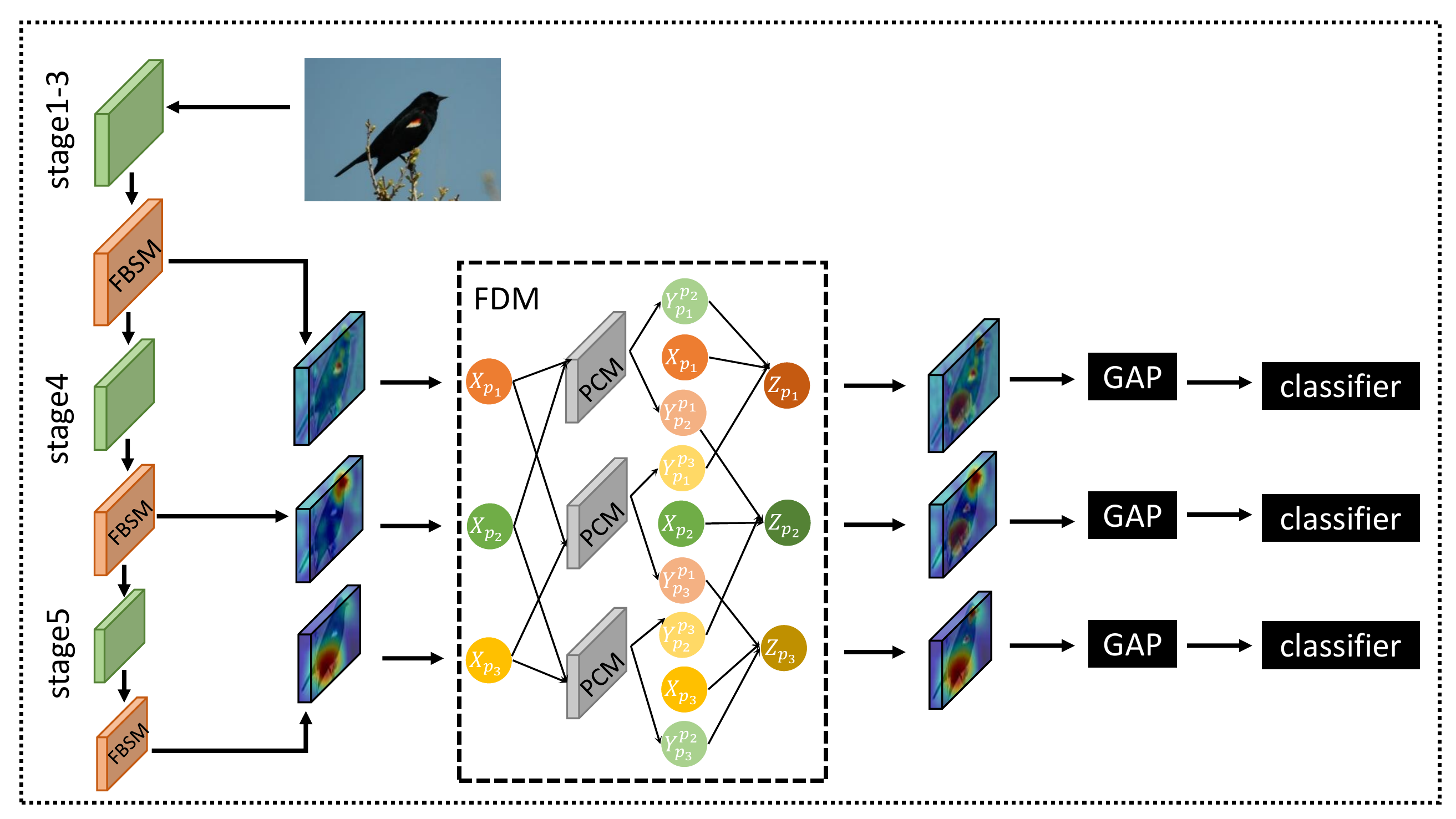}
		\caption{The overview of our method.}
		\label{fig:framework}
	\end{figure*}

	Finally, we jointly optimize FBSM and FDM as shown in Fig. \ref{fig:framework}.
	Our method does not need bounding boxes/part annotations and state-of-the-art performances are reported on several standard benchmark datasets. Moreover, our model is lightweight and easy to train as it does not involve the multi-crop mechanism\cite{RA-CNN}\cite{DB-Net}\cite{FDL}.
	
	Our contributions are summarized as follows:
	\begin{itemize}
		\item We propose a feature boosting and suppression module, which can explicitly force the network to focus on multiple discriminative parts.
		\item We propose a feature diversification module, which can model part interaction and diversify each part-specific representation.
	\end{itemize}
	\section{related work}
	Below, we review the most representative methods related to our method.
	\subsection{Fine-Grained Feature Learning}
	Ding et al.\cite{S3N} proposed sparse selective sampling learning to obtain both discriminative and complementary regions. Sun et al.\cite{MAMC} proposed a one-squeeze multi-excitation module to learn multiple parts, then applied a multi-attention multi-class constraint on these parts. Zhang et al.\cite{API-NET} proposed to discover contrastive clues by comparing image pairs. Yang et al.\cite{NTS} introduced a navigator-teacher-scrutinizer
	network to obtain discriminative regions. Luo et al.\cite{CrossX} proposed Cross-X learning to explore the relationships between different images and different layers. Gao et al.\cite{CIN} proposed to model channel interaction to capture subtle differences. Li et al.\cite{iSQRT-COV} proposed to capture the discrimination by matrix square root normalization and introduced an iterative method for fast end-to-end training. 
	Shi et al. \cite{Beyond-att} removed confusable features from the distinguishable parts to facilitate fine-grained classification. He et al. \cite{Progressive-Att} proposed progressive attention to localize parts at different scales.
	Our method utilizes feature boosting and suppression to learn different part representations in an explicit way, which is significantly different from previous methods. 
	\subsection{Feature Fusion}
	FPN\cite{FPN} and SSD\cite{SSD} aggregating feature maps from different layers have achieved great success in the object detection field. However, they use element-wise addition as the aggregation operation, making the capabilities of these methods still limited.  Wang el at.\cite{Non-Local} proposed a non-local operation that computes the response at a spatial position as a weighted sum of the features at all positions in the feature maps. SG-Net\cite{SG-Net} utilized the non-local operation to fuse feature maps from different layers. CIN \cite{CIN} adopted the non-local operation to mine semantically complementary information from different feature channels. 
	Our FDM is similar with \cite{SG-Net} and \cite{CIN}, but there are essential differences: (1) SG-Net tends to explore positive correlations to capture long-range dependencies, while FDM tends to explore negative correlations to diversify the feature representation. (2) CIN mines complementary information along channel dimension
	whereas FDM along the spatial dimension.
	
	\section{methodology}
	\begin{figure}
		\includegraphics[width=\linewidth]{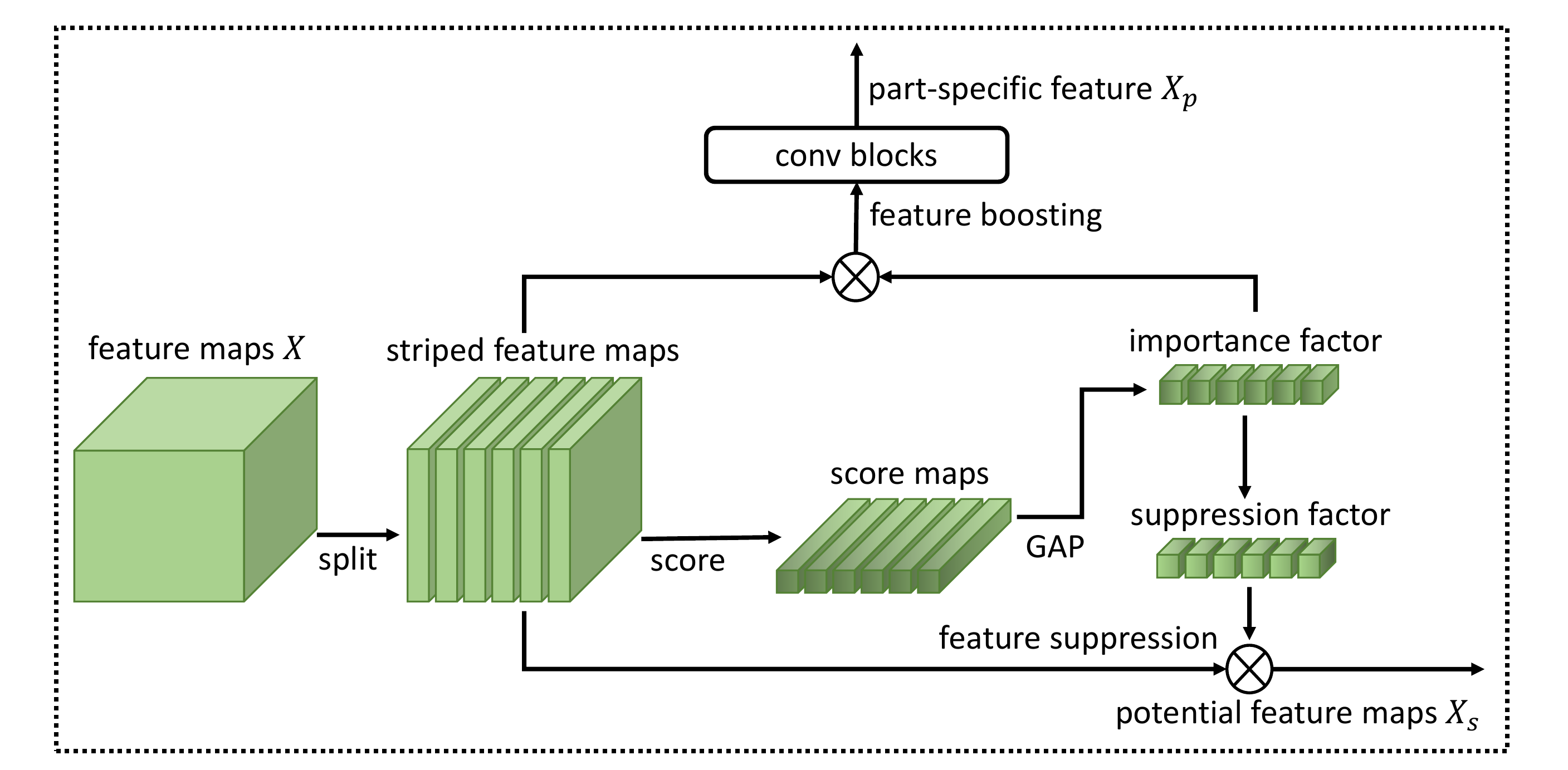}
		\caption{The diagram of the FBSM.}
		\label{fig:FBSM}
	\end{figure}
	In this section, we will detail the proposed method. An overview of the framework is shown in Fig. \ref{fig:framework}. Our model consists of two lightweight modules: (1) A feature boosting and suppression module (FBSM) aiming at 
	learning multiple discriminative part-specific representations as different as possible. 
	(2) A feature diversification module (FDM) aiming at modeling part interaction to enhance each part-specific representation.
	\subsection{Feature Boosting and Suppression Module}
	Given feature maps $X\in R^{C\times W\times H}$ from a specific layer, where $C$, $W$, $H$ represents the number of channels, width, and height respectively. 
	Inspired by \cite{Part-Pool}, we simply split $X$ evenly into $k$ parts along width dimension and 
	denote each striped parts as $X_{(i)}\in R^{C\times (W/k)\times H}$, $i\in [1,k]$. Then we employ a $1\times 1$ convolution $\phi$ to explore the importance of each part:
	\begin{equation}
		A_{(i)} = \text{Relu}(\phi(X_{(i)}))\in R^{1\times (W/k)\times H}
	\end{equation}
	The nonlinear function Relu\cite{Relu} is applied to remove the negative activations.
	$\phi$ is shared among different striped parts and acts as a grader. We then take the average of $A_{(i)}$ as the importance factor $b_{i}'$ for $X_{(i)}$, i.e.,
	\begin{equation}
		b_{i}' = \text{GAP}(A_{(i)})\in R
	\end{equation}
	where GAP denotes global average pooling. We use softmax to normalize $B'=(b'_1, \cdots, b'_k)^T$:
	\begin{equation}
		b_i = \dfrac{\text{exp}(b'_i)}{\sum_{j\in [1,k]}{\text{exp}(b'_j)}}
	\end{equation}
	With the normalized importance factors $B=(b_1,\cdots,b_k)^T$, the most salient part can be determined immediately. We then obtain the boosting feature $X_b$ by boosting the most salient part:
	\begin{equation}
		X_{b} = X + \alpha*(B\otimes X)
		\label{eq:boost}
	\end{equation}
	where $\alpha$ is a hyper-parameter, which controls the extent of boosting, $\otimes$ denotes element-wise multiplication.
	A convolutional layer $h$ is applied on $X_b$ to get a part-specific representation $X_{p}$:
	\begin{equation}
		X_{p} = h(X_b)
	\end{equation}
	By suppressing the most striped part, we can obtain the suppression feature $X_s$:
	\begin{equation}
		X_s = S\otimes X
	\end{equation}
	\begin{equation}
		s_i = 
		\begin{cases}
			1 - \beta, &  \text{if }b_i = \text{max}(B) \\
			1,     &  \text{otherwise}
		\end{cases}
		\label{eq:suppress}
	\end{equation}
	where $S=(s_1,\cdots, s_k)^T$, $\beta$ is a hyper-parameter, which controls the extent of suppressing.
	
	In short, the functionality of FBSM can be expressed as: $\text{FBSM}(X) = (X_{p}, X_{s})$. Given feature maps $X$, FBSM outputs part-specific feature $X_p$ and potential feature maps $X_s$. Since $X_s$ suppresses the most salient part in current stage, other potential parts will stand out after feeding $X_s$ into the following stage.
	A diagram of the FBSM is shown in Fig. \ref{fig:FBSM}.
	
	\subsection{Feature Diversification Module}
	As learning discriminative and diverse feature plays a key role in FGVC\cite{MA-CNN}\cite{MAMC}\cite{DB-Net}, we propose a feature diversification module, which enhances each part-specific feature
	by aggregating complementary information mined from other part-specific representations.
	
\begin{figure}
		\includegraphics[width=\linewidth]{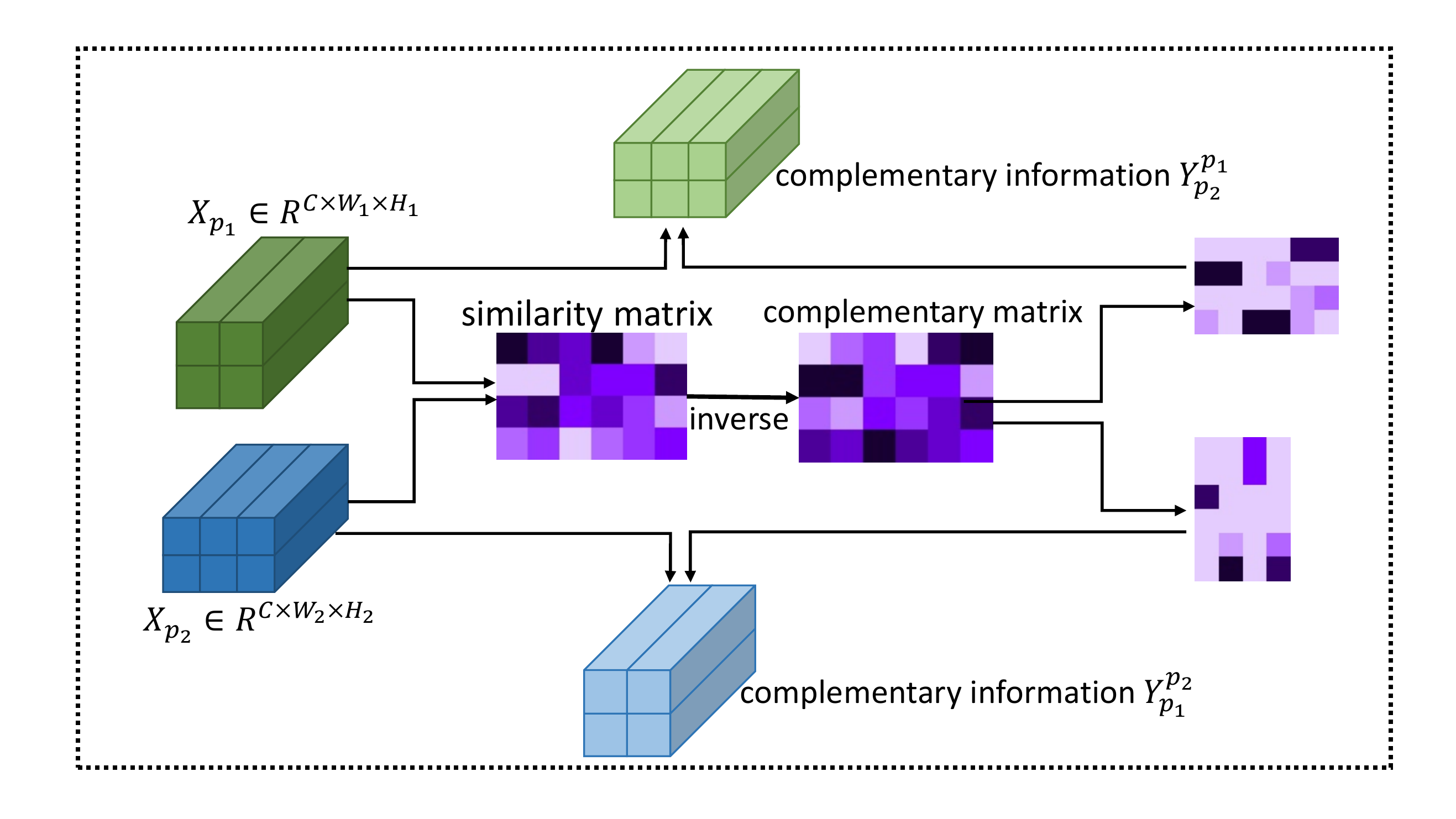}
		\caption{The diagram of the PCM.}
		\label{fig:PCM}
	\end{figure}

	We first discuss how two part-specific features diversify each other with the pairwise complement module (PCM). A simple illustration of PCM is shown in Fig. \ref{fig:PCM}.
	Without loss of generality, we denote two different part-specific features as $X_{p_1}\in R^{C\times W_1H_1}$ and $X_{p_2}\in R^{C\times W_2H_2}$, where $C$ denotes the number of channels, $W_1H_1$ and $W_2H_2$ denote their spatial size respectively.  We use subscript $p_i$ to denote that $X_{p_i}$ focuses on the $i^{th}$ part of the object and will omit the subscript when there is no ambiguity.
	We denote the feature vector at each spatial position along channel dimension as a pixel, i.e.,
	\begin{equation}
		\pixel(X, i) = (X_{1,i}, \cdots, X_{C,i})^T
	\end{equation}
	We first calculate the similarities between pixels in $X_{p_1}$ and pixels in $X_{p_2}$:
	\begin{equation}
		M = f(X_{p_1}, X_{p_2}), \quad f(X,Y)=X^TY
		\label{eq:M}
	\end{equation}
	Here, we use inner product to compute the similarity. The element $M_{i,j}$ represents the similarity of the $i^{th}$ pixel of $X_{p_1}$ and the $j^{th}$ pixel of $X_{p_2}$. 
	The lower the similarity of two pixels is, the more complementary they are, so we adopt $-M$ as the complementary matrix.
	Then we operate normalization on $-M$ row-wise and column-wise respectively:
	\begin{equation}
		A_{p_1}^{p_2} = \text{softmax}(-M^T)\in[0,1]^{W_2H_2\times W_1H_1}
		\label{eq:A1}
	\end{equation}
	\begin{equation}
		A_{p_2}^{p_1} = \text{softmax}(-M)\in[0,1]^{W_1H_1\times W_2H_2}
		\label{eq:A2}
	\end{equation}
	where softmax is performed column-wise. Then we can get the complementary information:
	\begin{equation}
		Y_{p_1}^{p_2} = X_{p_2}A_{p_1}^{p_2}\in R^{C\times W_1H_1}
		\label{eq:Y1}
	\end{equation}
	\begin{equation}
		Y_{p_2}^{p_1} = X_{p_1}A_{p_2}^{p_1}\in R^{C\times W_2H_2}
		\label{eq:Y2}
	\end{equation}
	where $Y_{p_i}^{p_j}$ denotes the complementary information of $X_{p_i}$ relative to $X_{p_j}$.
	It is worth noting that each pixel of $Y_{p_1}^{p_2}$ can be written as:
	\begin{equation}
		\pixel(Y_{p_1}^{p_2}, i)= \sum_{j\in[1, W_2H_2]}{(A_{p_1}^{p_2})_{i,j}}*\pixel(X_{p_2},j)
	\end{equation}
	i.e., each pixel of $Y_{p_1}^{p_2}$ takes all pixels of $X_{p_2}$ as references, and the higher the complementarity between $\pixel(X_{p_1}, i)$ and $\pixel(X_{p_2},j)$ is, the greater the contribution of $\pixel(X_{p_2},j)$ to $\pixel(Y_{p_1}^{p_2}, i)$ is.
	In this way, every pixel in these two part-specific features can mine semantically complementary information from each other.
	
	Now we discuss the general case. Formally, given a collection of part-specific features $P = \{X_{p_1}, X_{p_2}, X_{p_3}\cdots, X_{p_n}\}$, the complementary information of $X_{p_i}$ is:
	\begin{equation}
		Y_{p_i} = \sum_{X_{p_j}\in P\wedge i\neq j}Y_{p_i}^{p_j}
	\end{equation}
	where $Y_{p_i}^{p_j}$ can be obtained by applying $X_{p_i}$ and $X_{p_j}$ on \eqref{eq:M}, \eqref{eq:A1}, and \eqref{eq:Y1}. In practice, we can compute $Y_{p_i}^{p_j}$ and $Y_{p_j}^{p_i}$ simultaneously as shown in Fig. \ref{fig:PCM}.
	Then we get the enhanced part-specific feature:
	\begin{equation}
		Z_{p_i} = X_{p_i} + \gamma*Y_{p_i}
	\end{equation}
	where $\gamma$ is a hyper-parameter, which controls the extent of diversification.
	\subsection{Network Design}
	Our method can be easily implemented on various convolutional neural networks. 
	As shown in Fig. \ref{fig:framework}, we take Resnet\cite{ResNet} as an example.
	The feature extractor of Resnet has five stages and the spatial size of feature maps is halved after each stage. Considering that the deep layers have more semantic information, we plug FBSMs into the end of $stage_3$, $stage_4$, $stage_5$. 
	The different part-specific representations generated by FBSMs are fed into FDM to diversify each representation. Our method is highly customizable, it can adapt to different granularities of classification by adjusting the number of FBSMs directly.
	
	At training time, we compute the classification loss for each enhanced part-specific feature $Z_{p_i}$:
	\begin{align}
		L_{cls}^i = -y^T\text{log}(p_i), \quad
		p_i = \text{softmax}(\text{cls}_i(Z_{p_i}))
	\end{align}
	where $y$ is the ground-truth label of the input image and represented by one-hot vector, $\text{cls}_i$ is a classifier for the $i^{th}$ part, $p_i\in R^N$ is the prediction score vector, $N$ is the number of object categories.
	The final optimization objective is:
	\begin{equation}
		L = \sum_{i=1}^T L_{cls}^{i}
	\end{equation}
	where $T=3$ is the number of enhanced part-specific features. 
	At inference time, we take the average of prediction scores for all enhanced part-specific features as the final prediction result.
	\section{experiments}
	\begin{table}[t]
		\caption{Four fine-grained datasets commonly used in FGVC.}
		\label{table:dataset}
		\centering
		\begin{tabular}{ccccc}
			\hline
			Dataset       & Name     & \#Class & \#Train & \#Test \\ \hline
			CUB-200-2011  & Bird     & 200     & 5994   & 5794 \\
			FGVC-Aircraft & Aircraft & 100     & 6667   & 3333 \\
			Stanford Cars & Car      & 196     & 8144   & 8041 \\
			Stanford Dogs & Dog      & 120     & 12000  & 8580  \\ \hline
		\end{tabular}
	\end{table}
	\begin{table*}
		\caption{Comparison with state-of-the-art methods on four fine-grained benchmark datasets. ``-" means the result is not mentioned in the relevant paper.}
		\label{table:COMP}
		\centering
		\begin{tabular}{ccccccc}
			\hline
			\textbf{Methods}       & \textbf{Backbone}     
			&\textbf{1-Stage} & \textbf{CUB-200-2011} & \textbf{FGVC-Aircraft} 
			& \textbf{Stanford Cars} & \textbf{Stanford Dogs} \\ \hline
			DeepLAC       & VGG   &\unchecked       & 80.3 & - &- &- \\
			Part-RCNN     & VGG   &\unchecked       & 81.6 & - &- &- \\
			RA-CNN        & VGG   &\unchecked     & 85.3   & 88.1 & 92.5 & 87.3 \\
			MA-CNN        & VGG   &\checked       & 86.5   & 89.9 & 92.8 & -  \\
			\hline
			MAMC          & Resnet50   &\checked       & 86.2   &- &92.8 & 84.8 \\
			NTS           & Resnet50   &\unchecked     & 87.5   & 91.4 & 93.3 &- \\
			API-Net       & Resnet50   &\checked       & 87.7   & 93.0 & \textbf{94.8} &88.3\\
			Cross-X       & Resnet50   &\checked       & 87.7   & 92.6 & 94.5 & \textbf{88.9}\\
			DCL           & Resnet50   &\checked       & 87.8   & 93.0 & 94.5 & -\\
			DTB-Net       & Resnet50   &\checked      & 87.5   & 91.2 & 94.1 & -\\
			CIN           & Resnet50   &\checked       & 87.5   & 92.6 & 94.1 & -\\
			LIO           & Resnet50   &\checked       & 88.0   & 92.7 & 94.5 & -\\
			ISQRT-COV     & Resnet50   &\checked       & 88.1   & 90.0 & 92.8 & - \\
			MGE-CNN       & Resnet50   &\unchecked     & 88.5   &-     & 93.9 & - \\
			S3N           & Resnet50   &\unchecked     & 88.5   & 92.8 & 94.7 &-  \\ 
			FDL          & Resnet50   &\unchecked  & 88.6   & \textbf{93.4} & 94.3 &85.0 \\
			Ours          &Resnet50 &\checked    & \textbf{89.3} & 92.7 & 94.4 & 88.2\\ \hline
			
			MAMC          & Resnet101   &\checked       & 86.5   &-  &93.0  &85.2\\
			DTB-Net       &Resnet101   &\checked       & 88.1   &91.6 &94.5 &-\\
			CIN           & Resnet101   &\checked       & 88.1   &92.8 &94.5 &-\\
			API-Net       & Resnet101   &\checked       & 88.6   &\textbf{93.4} &94.9 &\textbf{90.3}\\
			ISQRT-COV     & Resnet101   &\checked       & 88.7   &91.4 &93.3 &-\\
			
			MGE-CNN       & Resnet101   &\unchecked     & 89.4   & - &93.6 & - \\
			Ours &Resnet101 &\checked    & \textbf{89.5}&93.1 &\textbf{95.0} &89.4\\ \hline
			FDL &Densenet161 &\checked &89.1 &91.3 &94.0 &84.9  \\
			Ours&Densenet161 &\checked    & \textbf{89.8}&\textbf{93.2} &\textbf{94.5} &\textbf{88.1}\\ \hline
		\end{tabular}
	\end{table*}
	\subsection{Datasets and Baselines}
	We evaluate our model on four commonly used datasets: CUB-200-2011\cite{CUB}, FGVC-Aircraft\cite{CRAFT}, Stanford Cars\cite{CAR}, Stanford Dogs\cite{DOG}. The details of each dataset can be found in Table \ref{table:dataset}. We compare our model with following baselines due to their state-of-the-art results. All baselines are listed as follows:
	\begin{itemize}
		\item \textbf{Part-RCNN}\cite{Part-RCNN}: proposes geometric constraints on mined semantic parts to normalize the pose.
		\item \textbf{DeepLAC}\cite{Deep-LAC}: integrates part localization, part alignment, and classification in one deep neural network.
		\item \textbf{S3N}\cite{S3N}: learns to mine discriminative and complementary parts to enhance the feature learning.
		\item \textbf{API-Net}\cite{API-NET}: proposes an attentive pairwise interaction network to identify differences by comparing image pairs.
		\item \textbf{NTS}\cite{NTS}: guides region proposal network by forcing the consistency between informativeness
		of the regions and their probabilities being ground-truth class.
		\item \textbf{MGE-CNN}\cite{MGE}: learns a mixture of granularity-specific experts to capture granularity-specific parts.
		\item \textbf{DCL}\cite{DCL}: learns to destruct and construct the image to acquire the expert knowledge.
		\item \textbf{MAMC}\cite{MAMC}: applies the multi-attention multi-class constraint in a
		metric learning framework to mine parts.
		\item \textbf{MA-CNN}\cite{MA-CNN}: makes part mining and fine-grained features learning in a mutual reinforced way.
		\item \textbf{RA-CNN}\cite{RA-CNN}: learns discriminative region attention at multiple scales recursively.
		\item \textbf{ISQRT-COV}\cite{BCNN}: utilizes bilinear information to model pairwise interaction.
		\item \textbf{DBT-Net}\cite{DBT-Net}: performs bilinear transformation on each semantically consistent channel group to model high-order information.
		\item \textbf{Cross-X}\cite{CrossX}: proposes to learn multi-scale feature representation between different layers and different images.
		\item \textbf{CIN}\cite{CIN}: models channel interaction to mine semantically complementary information.
		\item \textbf{FDL}\cite{FDL}: proposes to enhance discriminative region attention by filtration and distillation learning.
		\item \textbf{LIO}\cite{LIO}: proposes to model internal structure of
		the object to enhance feature learning.
	\end{itemize}
	\begin{figure*} 
		\centering
		\includegraphics[width=\linewidth]{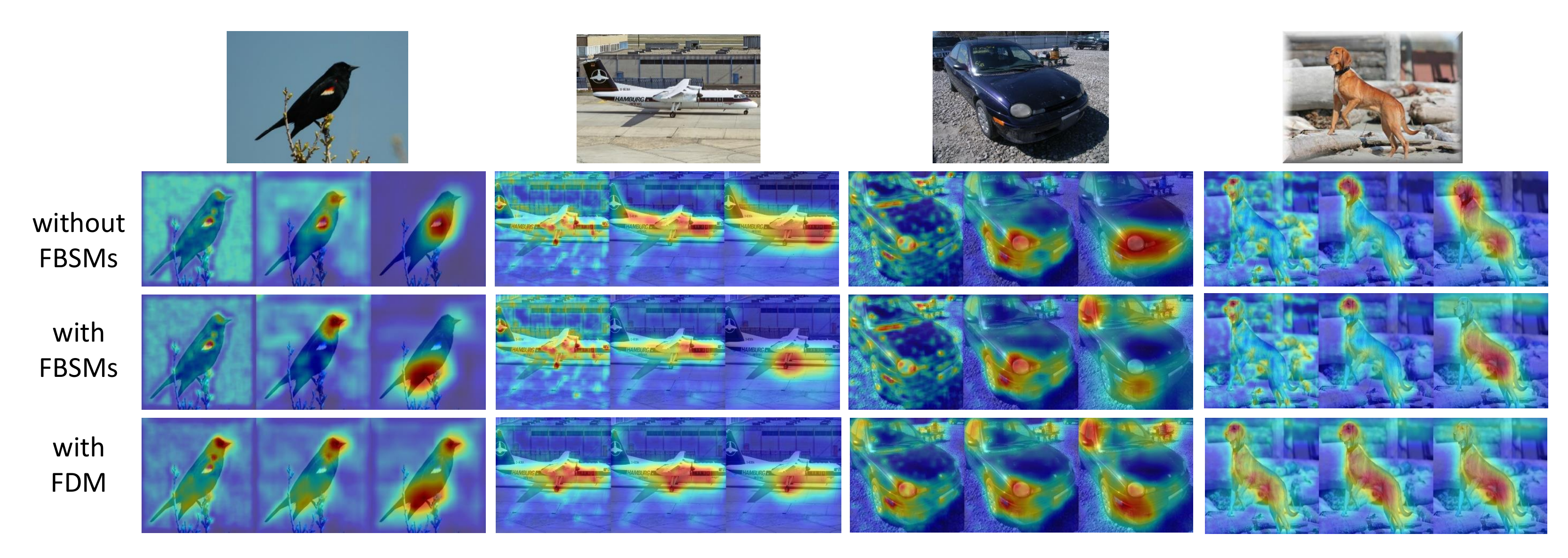}
		\caption{Visualization of the activation maps at different stages without FBSMs, with FBSMs, and with FDM on four benchmark datasets.}
		\label{fig:vis}
	\end{figure*}
	\subsection{Implementation Details}
	We validate the performance of our method on Resnet50, Resnet101\cite{ResNet} and Densenet161\cite{DenseNet}, which are all pre-trained on the ImageNet dataset\cite{ImageNet}. We insert FBSMs at the end of $stage_3$, $stage_4$ and $stage_5$.
	During training, the input images are resized to $550\times 550$ and randomly cropped to $448\times 448$. We apply random horizontal flips to augment the trainset.
	During testing, the input images are
	resized to $550\times 550$ and cropped from center into $448\times 448$. We set hyper-parameters $\alpha=0.5$, $\beta=0.5$ and $\gamma=1$.
	
	Our model is optimized	by Stochastic Gradient Descent with the momentum of 0.9, epoch number of 200,
	weight decay of 0.00001, mini-batch of 20. The learning rate of the backbone layers is set to 0.002, and the newly added layers are set to 0.02. The learning rate is adjusted by cosine anneal scheduler\cite{Warm}. We use PyTorch to implement our experiments. More implementation details can be referred to our code \url{https://github.com/chaomaer/FBSD}.
	\subsection{Comparison with State-of-the-Art}
	The top-1 classification accuracy on CUB-200-2011\cite{CUB}, FGVC-Aircraft\cite{CRAFT}, Stanford Cars\cite{CAR} and Stanford Dogs\cite{DOG} datasets are reported in Table \ref{table:COMP}.
	
	\textbf{Results on CUB-200-2011}: CUB-200-2011 is the most challenging benchmark in FGVC, our models based on Resnet50, Resnet101, and Densenet161 all achieve the best performances on this dataset. Compared with DeepLAC and Part-RCNN which use predefined bounding boxes/part annotations, our method is 9.0\%, 7.7\% higher than them. Compared with the two-stage methods: RA-CNN, NTS, MGE-CNN, S3N, and FDL, which all take the raw image as input at the first stage to explore informative regions and takes them as input at the second stage, our model is 4.0\%, 1.8\%, 0.8\%, 0.8\%, 0.7\% higher than them respectively. ISQRT-COV and DTB-Net explore high-order information to capture the subtle differences, our method outperforms them by large margins. Compared with API-Net and Cross-X, which both take image pairs as input and model the discrimination by part interaction, our model gets 1.6\% improvements. The accuracy of our method is 3.1\% higher than MAMC, which formulates part mining into a metric learning problem. Compared with MA-CNN, CIN, and LIO, our method is 2.8\%, 1.9\%, 1.3\% higher than them respectively. 
	DCL spots discriminative parts by diving into the destructed image, our method surpasses it by 1.5\%. Notably, our method based on Resnet50 outperforms almost all other methods based on Resnet101.
	
	\textbf{Results on FGVC-Aircraft}: Our method gets competitive results on this dataset. Compare with RA-CNN and MA-CNN, our method exceeds them by large margins. With Resnet50 backbone, our model is 2.7\%, 1.5\%, 1.3\%, 0.1\%, 0.1\% higher than ISQRT-COV, DTB-Net, NTS, Cross-X, and CIN respectively. 
	LIO that enhances feature learning by modeling object structure obtains the same result as our model. 
	Our model is 0.1\%, 0.3\%, 0.3\%, 0.7\% lower than S3N, DCL, API-Net and FDL. However, S3N and FDL are both two-stage methods whereas our method is one-stage and more efficient. DCL destructs the image to locate discriminative regions, but it is not easy to define what level of destruction is appropriate. API-Net needs to consider different pairwise image combinations and requires large computing resources.
	
	\textbf{Results on Stanford Cars}: Our method equipped with Resnet101 gets the best result on this dataset. Our method exceeds RA-CNN and MA-CNN which use VGG\cite{VGG} as backbone by large margins. With Resnet50 backbone, our method is higher than ISQRT-COV, MAMC, NTS, MEG-CNN, DTB-Net, CIN, and FDL but lower than DCL, LIO, Cross-X, S3N, and API-Net.
	We suspect that features extracted from shadow layers ($stage_3\& stage_4$) lack rich semantic information, which may cause degradation of recognition performance.
	When deepening the network and taking Resnet101 as the backbone, we obtain the best result of 95.0\%.
	
	\textbf{Results on Stanford Dogs}: Most previous methods do not report results on this dataset because of the computational complexity.
	Our method obtains a competitive result on this dataset and surpasses RA-CNN, MAMC, and FDL by large margins. Compared with Cross-X and API-Net which take image pairs as input, our method does not need to consider how to design a non-trivial data sampler to sample inter-class and intra-class image pairs\cite{Sampling}.
	
	In summary, due to the simplicity and effectiveness of our model, it scales well to all four benchmark datasets. With Resnet50 backbone, API-Net and Cross-X obtain the best result on Stanford Cars and Stanford Dogs respectively, but both get poor results on CUB-200-2011. FDL obtains the best result on FGVC-Aircraft but behaves inferiorly on 
	Stanford Dogs. Our model achieves the best result on CUB-200-2011 and relatively competitive results on the other three datasets.
	\begin{table}
		\caption{Ablation Studies on four benchmark datasets}
		\label{table:ablation}
		\centering
		\begin{tabular}{ccccc}
			\hline
			\textbf{Methods} &\textbf{Bird}  &\textbf{Aircraft} 
			&\textbf{Car} &\textbf{Dog}\\ \hline
			Resnet50 &85.5 &90.3 &89.8 &81.1 \\
			Resnet50+FBSM &88.9 &92.4 &94.0 &87.5 \\
			Resnet50+FBSM+FDM &89.3 &92.7 &94.4 &88.2 \\
			\hline
		\end{tabular}
	\end{table}
	\subsection{Ablation Studies}
	We perform ablation studies to understand the contributions of each proposed module. We make experiments on four datasets with Resnet50 as the backbone.  The results are reported in Table \ref{table:ablation}.

	\textbf{The effect of FBSM: }
	To obtain multiple discriminative part-specific feature representations, we insert FBSMs at the end of $stage_3$, $stage_4$ and $stage_5$ of Resnet50. With this module, the accuracy of Bird, Aircraft, Car, and Dog increased by 3.4\%, 2.1\%, 4.2\%, and 6.4\% respectively, which reflects the effectiveness of the FBSM.
	
	\textbf{The effect of FDM: }
	When introducing FDM into our approach to model part interaction, the classification results on Bird, Aircraft, Car, and Dog datasets increased by 0.4\%, 0.3\%, 0.4\%, and 0.7\% respectively, which indicates the effectiveness of the FDM.
	It is worth noting that FDM does not introduce additional learning parameters.
	\subsection{Visualization}
	We visualize the activation maps taken from Resnet50 without FBSMs, with FBSMs, and with FDM on four benchmark datasets. Specifically, the activation map is obtained by averaging the activation values across the channel dimension given feature maps.
	As shown in Fig. \ref{fig:vis}, for each raw image sampled from four datasets, the activation maps at the first to third columns correspond to the third to fifth stages of Resnet50 respectively.
	We can observe that the network tends to focus on the most salient part without FBSMs and is forced to mine different parts when equipped with FBSMs.
	Taking the bird as an example, without FBSMs, the features at different stages all focus on the swing. When there are FBSMs,
	the features in $stage_3$ focus on the swing, the features in $stage_4$ focus on the head, and the features in $stage_5$ focus on the tail. 
	When introducing FDM, the features in all stages focus on the entire parts mined by different stages.
	The visualization experiments prove the capability of FBSMs for mining multiple different discriminative object parts and FDM for diversifying feature representation.
	\section{conclusion}
	In this paper, we propose to learn feature boosting, suppression, and diversification for fine-grained visual classification. Specifically, we introduce two lightweight modules: One is the feature boosting and suppression module which boosts the most salient part of the feature maps to obtain the part-specific feature and suppresses it to explicitly force following stages to mine other potential parts.
	The other is the feature diversification module which aggregates semantically complementary information from other object parts to each part-specific representation.
	The synergy between these two modules helps the network to learn more discriminative and diverse feature representations. Our method can be trained end-to-end and does not need bounding boxes/part annotations. The state-of-the-art results are obtained on several benchmark datasets and ablation studies further prove the effectiveness of each proposed module. In the future, we will investigate how to adaptively divide feature maps into suitable patches to boost and suppress, instead of simple striped parts.
	\section*{ACKNOWLEDGMENT}
	This work is supported by the National Key Research and Development Project under Grant No.SQ2020YFB1707600.		   
\input{conference_101719.bbl}

\end{document}

%% file: conference_101719.bbl

%% file: conference_101719.bbl
\begin{thebibliography}{10}
\providecommand{\url}[1]{#1}
\csname url@samestyle\endcsname
\providecommand{\newblock}{\relax}
\providecommand{\bibinfo}[2]{#2}
\providecommand{\BIBentrySTDinterwordspacing}{\spaceskip=0pt\relax}
\providecommand{\BIBentryALTinterwordstretchfactor}{4}
\providecommand{\BIBentryALTinterwordspacing}{\spaceskip=\fontdimen2\font plus
\BIBentryALTinterwordstretchfactor\fontdimen3\font minus
  \fontdimen4\font\relax}
\providecommand{\BIBforeignlanguage}[2]{{%
\expandafter\ifx\csname l@#1\endcsname\relax
\typeout{** WARNING: IEEEtran.bst: No hyphenation pattern has been}%
\typeout{** loaded for the language `#1'. Using the pattern for}%
\typeout{** the default language instead.}%
\else
\language=\csname l@#1\endcsname
\fi
#2}}
\providecommand{\BIBdecl}{\relax}
\BIBdecl

\bibitem{CUB}
P.~Welinder, S.~Branson, T.~Mita, C.~Wah, F.~Schroff, S.~Belongie, and
  P.~Perona, ``{Caltech-UCSD Birds 200},'' California Institute of Technology,
  Tech. Rep. CNS-TR-2010-001, 2010.

\bibitem{DOG}
A.~Khosla, N.~Jayadevaprakash, B.~Yao, and L.~Fei-Fei, ``Novel dataset for
  fine-grained image categorization,'' in \emph{First Workshop on Fine-Grained
  Visual Categorization, IEEE Conference on Computer Vision and Pattern
  Recognition}, Colorado Springs, CO, June 2011.

\bibitem{CRAFT}
S.~Maji, J.~Kannala, E.~Rahtu, M.~Blaschko, and A.~Vedaldi, ``Fine-grained
  visual classification of aircraft,'' Tech. Rep., 2013.

\bibitem{CAR}
J.~Krause, M.~Stark, J.~Deng, and L.~Fei-Fei, ``3d object representations for
  fine-grained categorization,'' in \emph{4th International IEEE Workshop on 3D
  Representation and Recognition (3dRR-13)}, Sydney, Australia, 2013.

\bibitem{facenet}
F.~Schroff, D.~Kalenichenko, and J.~Philbin, ``Facenet: A unified embedding for
  face recognition and clustering,'' in \emph{Proceedings of the IEEE
  conference on computer vision and pattern recognition}, 2015, pp. 815--823.

\bibitem{self-drive}
M.~Bojarski, D.~Del~Testa, D.~Dworakowski, B.~Firner, B.~Flepp, P.~Goyal, L.~D.
  Jackel, M.~Monfort, U.~Muller, J.~Zhang \emph{et~al.}, ``End to end learning
  for self-driving cars,'' \emph{arXiv preprint arXiv:1604.07316}, 2016.

\bibitem{Part-Pool}
Y.~Sun, L.~Zheng, Y.~Yang, Q.~Tian, and S.~Wang, ``Beyond part models: Person
  retrieval with refined part pooling (and a strong convolutional baseline),''
  in \emph{Proceedings of the European Conference on Computer Vision (ECCV)},
  2018, pp. 480--496.

\bibitem{Part-RCNN}
N.~Zhang, J.~Donahue, R.~Girshick, and T.~Darrell, ``Part-based r-cnns for
  fine-grained category detection,'' \emph{Lecture Notes in Computer Science},
  p. 834–849, 2014.

\bibitem{Deep-LAC}
D.~{Lin}, X.~{Shen}, C.~{Lu}, and J.~{Jia}, ``Deep lac: Deep localization,
  alignment and classification for fine-grained recognition,'' in \emph{2015
  IEEE Conference on Computer Vision and Pattern Recognition (CVPR)}, 2015, pp.
  1666--1674.

\bibitem{Part-SCNN}
S.~Huang, Z.~Xu, D.~Tao, and Y.~Zhang, ``Part-stacked cnn for fine-grained
  visual categorization,'' \emph{2016 IEEE Conference on Computer Vision and
  Pattern Recognition (CVPR)}, Jun 2016.

\bibitem{Pose-Norm}
S.~Branson, G.~V. Horn, S.~Belongie, and P.~Perona, ``Bird species
  categorization using pose normalized deep convolutional nets,'' 2014.

\bibitem{RA-CNN}
J.~{Fu}, H.~{Zheng}, and T.~{Mei}, ``Look closer to see better: Recurrent
  attention convolutional neural network for fine-grained image recognition,''
  in \emph{2017 IEEE Conference on Computer Vision and Pattern Recognition
  (CVPR)}, 2017, pp. 4476--4484.

\bibitem{FDL}
C.~Liu, H.~Xie, Z.~Zha, L.~Ma, L.~Yu, and Y.~Zhang, ``Filtration and
  distillation: Enhancing region attention for fine-grained visual
  categorization,'' in \emph{Proceedings of the AAAI Conference on Artificial
  Intelligence}.\hskip 1em plus 0.5em minus 0.4em\relax {AAAI} Press, 2020, pp.
  11\,555--11\,562.

\bibitem{NTS}
Z.~Yang, T.~Luo, D.~Wang, Z.~Hu, J.~Gao, and L.~Wang, ``Learning to navigate
  for fine-grained classification,'' \emph{Lecture Notes in Computer Science},
  p. 438–454, 2018.

\bibitem{MGE}
L.~{Zhang}, S.~{Huang}, W.~{Liu}, and D.~{Tao}, ``Learning a mixture of
  granularity-specific experts for fine-grained categorization,'' in \emph{2019
  IEEE/CVF International Conference on Computer Vision (ICCV)}, 2019, pp.
  8330--8339.

\bibitem{S3N}
Y.~{Ding}, Y.~{Zhou}, Y.~{Zhu}, Q.~{Ye}, and J.~{Jiao}, ``Selective sparse
  sampling for fine-grained image recognition,'' in \emph{2019 IEEE/CVF
  International Conference on Computer Vision (ICCV)}, 2019, pp. 6598--6607.

\bibitem{BCNN}
T.-Y. Lin, A.~RoyChowdhury, and S.~Maji, ``Bilinear cnn models for fine-grained
  visual recognition,'' in \emph{Proceedings of the 2015 IEEE International
  Conference on Computer Vision (ICCV)}, ser. ICCV ’15.\hskip 1em plus 0.5em
  minus 0.4em\relax USA: IEEE Computer Society, 2015, p. 1449–1457.

\bibitem{Compact-BCNN}
Y.~Gao, O.~Beijbom, N.~Zhang, and T.~Darrell, ``Compact bilinear pooling,''
  \emph{2016 IEEE Conference on Computer Vision and Pattern Recognition
  (CVPR)}, Jun 2016.

\bibitem{iSQRT-COV}
P.~Li, J.~Xie, Q.~Wang, and Z.~Gao, ``Towards faster training of global
  covariance pooling networks by iterative matrix square root normalization,''
  in \emph{IEEE Int. Conf. on Computer Vision and Pattern Recognition (CVPR)},
  June 2018.

\bibitem{HCNN}
S.~{Cai}, W.~{Zuo}, and L.~{Zhang}, ``Higher-order integration of hierarchical
  convolutional activations for fine-grained visual categorization,'' in
  \emph{2017 IEEE International Conference on Computer Vision (ICCV)}, 2017,
  pp. 511--520.

\bibitem{Low-Rank}
S.~Kong and C.~Fowlkes, ``Low-rank bilinear pooling for fine-grained
  classification,'' in \emph{Proceedings of the IEEE conference on computer
  vision and pattern recognition}, 2017, pp. 365--374.

\bibitem{GA-CNN}
J.~Song and R.~Yang, ``Learning granularity-aware convolutional neural network
  for fine-grained visual classification,'' \emph{arXiv preprint
  arXiv:2103.02788}, 2021.

\bibitem{DB-Net}
G.~Sun, H.~Cholakkal, S.~Khan, F.~Khan, and L.~Shao, ``Fine-grained
  recognition: Accounting for subtle differences between similar classes,'' in
  \emph{Proceedings of the AAAI Conference on Artificial Intelligence},
  vol.~34, no.~07, 2020, pp. 12\,047--12\,054.

\bibitem{MAMC}
M.~Sun, Y.~Yuan, F.~Zhou, and E.~Ding, ``Multi-attention multi-class constraint
  for fine-grained image recognition,'' \emph{Lecture Notes in Computer
  Science}, p. 834–850, 2018.

\bibitem{API-NET}
P.~Zhuang, Y.~Wang, and Y.~Qiao, ``Learning attentive pairwise interaction for
  fine-grained classification,'' \emph{Proceedings of the AAAI Conference on
  Artificial Intelligence}, vol.~34, no.~07, p. 13130–13137, Apr 2020.

\bibitem{CrossX}
W.~Luo, X.~Yang, X.~Mo, Y.~Lu, L.~S. Davis, and S.-N. Lim, ``Cross-x learning
  for fine-grained visual categorization,'' in \emph{ICCV}, 2019.

\bibitem{CIN}
Y.~Gao, X.~Han, X.~Wang, W.~Huang, and M.~Scott, ``Channel interaction networks
  for fine-grained image categorization.'' in \emph{AAAI}, 2020, pp.
  10\,818--10\,825.

\bibitem{Beyond-att}
X.~Shi, L.~Xu, P.~Wang, Y.~Gao, H.~Jian, and W.~Liu, ``Beyond the attention:
  Distinguish the discriminative and confusable features for fine-grained image
  classification,'' in \emph{Proceedings of the 28th ACM International
  Conference on Multimedia}, 2020, pp. 601--609.

\bibitem{Progressive-Att}
H.~Zheng, J.~Fu, Z.-J. Zha, J.~Luo, and T.~Mei, ``Learning rich part
  hierarchies with progressive attention networks for fine-grained image
  recognition,'' \emph{IEEE Transactions on Image Processing}, vol.~29, pp.
  476--488, 2019.

\bibitem{FPN}
T.-Y. Lin, P.~Dollar, R.~Girshick, K.~He, B.~Hariharan, and S.~Belongie,
  ``Feature pyramid networks for object detection,'' \emph{2017 IEEE Conference
  on Computer Vision and Pattern Recognition (CVPR)}, Jul 2017.

\bibitem{SSD}
W.~Liu, D.~Anguelov, D.~Erhan, C.~Szegedy, S.~Reed, C.-Y. Fu, and A.~C. Berg,
  ``Ssd: Single shot multibox detector,'' \emph{Lecture Notes in Computer
  Science}, p. 21–37, 2016.

\bibitem{Non-Local}
X.~Wang, R.~Girshick, A.~Gupta, and K.~He, ``Non-local neural networks,'' in
  \emph{Proceedings of the IEEE conference on computer vision and pattern
  recognition}, 2018, pp. 7794--7803.

\bibitem{SG-Net}
X.~Chen, C.~Fu, Y.~Zhao, F.~Zheng, J.~Song, R.~Ji, and Y.~Yang,
  ``Salience-guided cascaded suppression network for person
  re-identification,'' in \emph{Proceedings of the IEEE/CVF Conference on
  Computer Vision and Pattern Recognition}, 2020, pp. 3300--3310.

\bibitem{Relu}
A.~F. Agarap, ``Deep learning using rectified linear units (relu),''
  \emph{arXiv preprint arXiv:1803.08375}, 2018.

\bibitem{MA-CNN}
H.~Zheng, J.~Fu, T.~Mei, and J.~Luo, ``Learning multi-attention convolutional
  neural network for fine-grained image recognition,'' in \emph{Proceedings of
  the IEEE international conference on computer vision}, 2017, pp. 5209--5217.

\bibitem{ResNet}
K.~He, X.~Zhang, S.~Ren, and J.~Sun, ``Deep residual learning for image
  recognition,'' \emph{2016 IEEE Conference on Computer Vision and Pattern
  Recognition (CVPR)}, Jun 2016.

\bibitem{DCL}
Y.~Chen, Y.~Bai, W.~Zhang, and T.~Mei, ``Destruction and construction learning
  for fine-grained image recognition,'' in \emph{Proceedings of the IEEE
  Conference on Computer Vision and Pattern Recognition}, 2019, pp. 5157--5166.

\bibitem{DBT-Net}
H.~Zheng, J.~Fu, Z.-J. Zha, and J.~Luo, ``Learning deep bilinear transformation
  for fine-grained image representation,'' in \emph{Advances in Neural
  Information Processing Systems}, 2019, pp. 4277--4286.

\bibitem{LIO}
M.~Zhou, Y.~Bai, W.~Zhang, T.~Zhao, and T.~Mei, ``Look-into-object:
  Self-supervised structure modeling for object recognition,'' in
  \emph{Proceedings of the IEEE/CVF Conference on Computer Vision and Pattern
  Recognition}, 2020, pp. 11\,774--11\,783.

\bibitem{DenseNet}
G.~{Huang}, Z.~{Liu}, L.~{Van Der Maaten}, and K.~Q. {Weinberger}, ``Densely
  connected convolutional networks,'' in \emph{2017 IEEE Conference on Computer
  Vision and Pattern Recognition (CVPR)}, 2017, pp. 2261--2269.

\bibitem{ImageNet}
J.~Deng, W.~Dong, R.~Socher, L.-J. Li, K.~Li, and L.~Fei-Fei, ``{ImageNet: A
  Large-Scale Hierarchical Image Database},'' in \emph{CVPR09}, 2009.

\bibitem{Warm}
I.~Loshchilov and F.~Hutter, ``Sgdr: Stochastic gradient descent with warm
  restarts,'' \emph{arXiv preprint arXiv:1608.03983}, 2016.

\bibitem{VGG}
K.~Simonyan and A.~Zisserman, ``Very deep convolutional networks for
  large-scale image recognition,'' \emph{arXiv preprint arXiv:1409.1556}, 2014.

\bibitem{Sampling}
C.-Y. Wu, R.~Manmatha, A.~J. Smola, and P.~Krahenbuhl, ``Sampling matters in
  deep embedding learning,'' in \emph{Proceedings of the IEEE International
  Conference on Computer Vision}, 2017, pp. 2840--2848.

\end{thebibliography}
